%% file: main_arxiv.tex
\documentclass{article}
\usepackage{microtype}
\usepackage{graphicx}
\usepackage{subcaption}
\usepackage{booktabs} 
\usepackage{hyperref}

\usepackage{algorithm}
\usepackage{algorithmic}

\usepackage{xspace}
\usepackage{booktabs}
\usepackage[table]{xcolor} 
\usepackage{caption}

\usepackage[preprint]{icml2026}

\usepackage{amsmath}
\usepackage{amssymb}
\usepackage{mathtools}
\usepackage{amsthm}
\usepackage[capitalize,noabbrev]{cleveref}

\usepackage{booktabs}
\theoremstyle{plain}

\theoremstyle{definition}

\theoremstyle{remark}

\usepackage{makecell} 
\usepackage{tikz}
\usepackage{subcaption}
\usepackage{listings}
\usepackage{xcolor}

\newcommand{\methodacro}{\textsc{ECS}\xspace}

\usepackage[textsize=tiny]{todonotes}

\begin{document}

\twocolumn[
  \icmltitle{Evolutionary Context Search for Automated Skill Acquisition}
  \icmlsetsymbol{equal}{*}
  \begin{icmlauthorlist}
    \icmlauthor{Qi Sun}{sakana,tokyo}
    \icmlauthor{Stefan Nielsen}{sakana}
    \icmlauthor{Rio Yokota}{tokyo}
    \icmlauthor{Yujin Tang}{sakana}
  \end{icmlauthorlist}

  \icmlaffiliation{sakana}{Sakana AI, Japan}
  \icmlaffiliation{tokyo}{Institute of Science Tokyo, Japan}
  \icmlcorrespondingauthor{Qi Sun}{qisun@sakana.ai}
  
  \icmlkeywords{Machine Learning, ICML}

  \vskip 0.3in
]

\printAffiliationsAndNotice{}  

\begin{abstract}
Large Language Models cannot reliably acquire new knowledge post-deployment---even when relevant text resources exist, models fail to transform them into actionable knowledge without retraining.
Retrieval-Augmented Generation attempts to bridge this gap by surfacing relevant documents at inference time, yet similarity-based retrieval often fails to identify context that actually improves task performance.
We introduce Evolutionary Context Search (ECS), an evolutionary method that searches context combinations using accuracy on a small development set, requiring only inference calls without weight updates.
ECS moves beyond semantic similarity to discover non-obvious context pairings that significantly boost performance.
Our empirical results show that ECS improves BackendBench by 27\% and $\tau$-bench airline by 7\%.
The evolved contexts are model-agnostic, as those evolved with Gemini-3-Flash transfer effectively to Claude Sonnet and DeepSeek.
This suggests that ECS opens a path toward automated context discovery for skill acquisition---an efficient alternative to manual prompt engineering or costly fine-tuning.
\end{abstract}

\input{Sections/introduction}
\input{Sections/related_works}

\input{Sections/method}
\input{Sections/experiments}
\input{Sections/analysis}
\input{Sections/conclusion}

\section*{Impact Statement}

This work introduces Evolutionary Context Search (ECS), a efficient framework that injecting knowledge from provided corpus into deployed LLM via context evolution.
By replacing resource-intensive fine-tuning with inference-time context search, our approach promotes sustainable AI development and democratizes access to domain adaptation.
Furthermore, ECS enhances the reliability of autonomous agents by discovering contexts that improve adherence to complex safety policies and operational constraints. 
Unlike opaque weight updates, the evolved contexts remain human-interpretable, fostering greater transparency in how models acquire and apply new knowledge in real-world deployment.

\bibliography{main}
\bibliographystyle{icml2026}

\input{Sections/appendix}

\end{document}

%% file: Sections/introduction.tex
\section{Introduction}

Updating the knowledge of Large Language Models (LLMs) to acquire new capabilities after the training cutoff remains a technical challenge \citep{onoe2023can, zhong2023mquake, yao2023editing, li2023unveiling}.
Domain-Specific Languages (DSLs) like CuTeDSL for GPU programming have comprehensive documentation, yet adapting LLMs to write code correctly in such high-resource but unseen languages cannot be done reliably \citep{kandpal2023large, gu2025effectiveness}. 
The core problem is not missing information, but effective methods to harness novel and diverse information sources to efficiently adapt the pretrained knowledge of an LLM to acquire the new target capability.

Existing approaches to skill acquisition incur substantial computational costs while struggling to obtain the required skill. 
Training-based methods, such as supervised finetuning (SFT) and reinforcement learning (RL) on curated data, are expensive due to their computational requirements, with additional engineering costs incurred by data collection and processing \citep{cottier2024rising}. 
Moreover, given post-training requires weight access, such methods are naturally inapplicable to frontier, closed-source models. 
Current in-context approaches offer only partial solutions to training-based methods. 
Retrieval-Augmented Generation (RAG) \citep{lewis2020retrieval, ram2023context, khandelwal2019generalization} can equip base models with new knowledge at test-time without necessitating weight access, but similarity-based retrieval often fails because queries tend to be verbose, contain irrelevant context, or not task-specific \citep{li2023large, petroni2020context, yoran2023making}.
More generally, RAG is highly sensitive to arbitrary context ordering and requires considerable human engineering effort \citep{akkiraju2024facts}, impairing the efficacy and viability of the technique.

\begin{figure*}[t]
    \centering
    \includegraphics[width=1.0\textwidth]{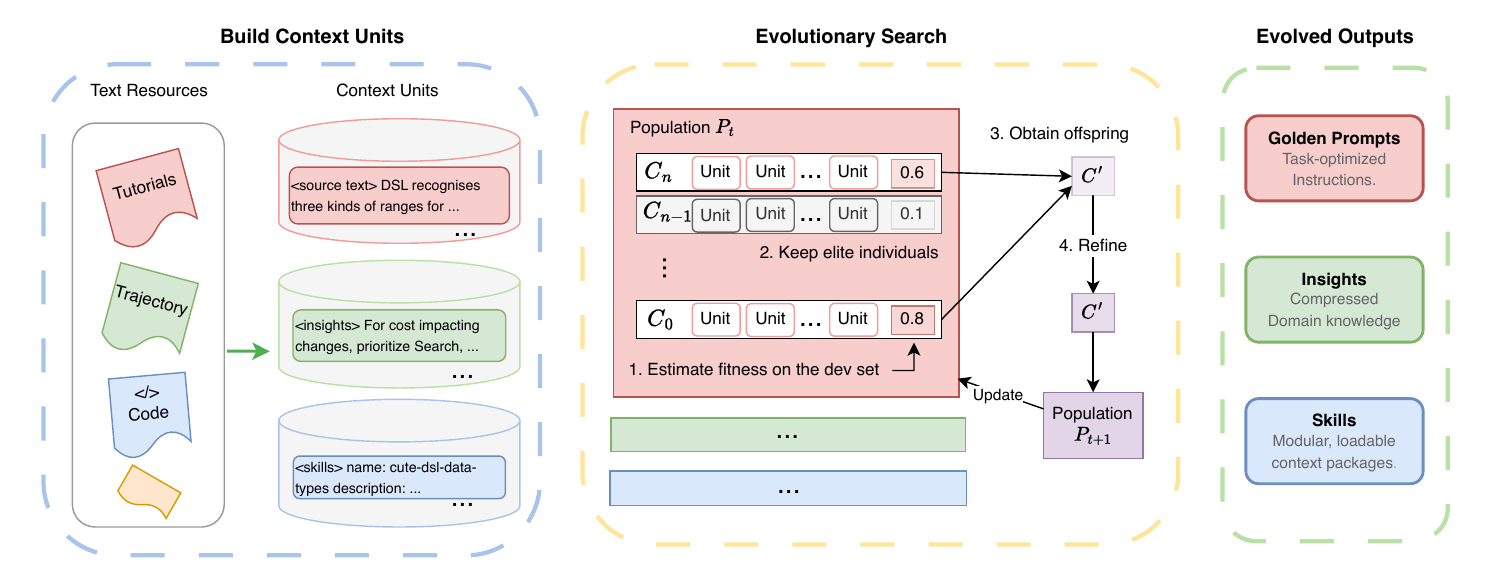}
    \caption{\textbf{Evolutionary Context Search.} Our method takes a population of text resources and evolves optimized contexts that confer the knowledge required to perform tasks in unseen domains. Each successive generation accumulates task-dependent, knowledge-rich information, effectively searching the corpora to obtain token-efficient contexts that enable novel skill acquisition in LLMs.}
    \label{fig:cover-image}
\end{figure*}

In this work, we leverage the viewpoint that text-based prompt augmentations offers a flexible and effective framework for updating an LLM's knowledge, and develop a search method for accumulating the required context that avoids the issues of retrieval entirely. 
Our core insight is that optimal prompt construction for novel skill acquisition ought to be an evolutionary search process.
Crucially, rather than relying on LLMs as evolutionary operators, we employ a simple genetic algorithm-style~\citep{katoch2021review} approach to evolve context combinations.
We find this process to be highly efficient, typically requiring as few as 5 iterations to converge on high-utility results.
This search process intuitively mirrors how humans learn new skills -- collection of relevant documentation and other resources, assessing the fitness of a given resource by how it improves our understanding, using such resources to discover followup works, and so on recursively until we accumulate the required contextual information to attain the new capability.


Building on these insights, we introduce Evolutionary Context Search (\methodacro) (Figure~ \ref{fig:cover-image}), a framework that extracts knowledge from provided corpora through fitness-guided evolution rather than similarity-based retrieval.
\methodacro takes a context-centric view: \methodacro converts text resources into context units and combines them into candidate contexts. 
These contexts are then evolved to improve performance on specific tasks. 
The evolved context can take multiple forms: raw documentation, condensed summaries/insights, or structured agent skills~\citep{anthropic2025skills}.
This approach transforms passive text resources into active teaching materials that progressively improve model performance. 
Our approach provides three connected benefits: 1) substantive performance gains beyond baseline methods as ECS evolves optimal contexts to provide base LLMs with the required new knowledge, 2) improved robustness to non-task-specific, verbose, or misleading queries, which markedly deteriorate the performance of retrieval-based methods, and 3) efficiency gains, as our method requires no weight or model access, or any training.

Our contributions are three-fold:
\begin{enumerate}
    \item We propose Evolutionary Context Search (\methodacro), a framework that treats context seleciton as an optimization problem to maximize skill and knowledge acquisition from external resources.
    \item We empirically demonstrate that \methodacro provides substantial improvements over baselines across diverse benchmarks, including $\tau^2$-bench and BackendBench.
    \item We show that the contexts discovered by \methodacro are model-agnostic and highly transferable, suggesting that \methodacro unlocks a new paradigm for injecting corpus knowledge into deployed LLMs. 
\end{enumerate}

%% file: Sections/related_works.tex
\section{Related Work}

\textbf{Knowledge updating.} Existing approaches to update an LLM’s knowledge to unseen information can be broken into two sets, gradient-based and in-context. Gradient-based approaches include supervised fine tuning and reinforcement learning from curated data, which incur persistent challenges due to catastrophic forgetting \citep{luo2025empirical, shi2025continual}, not to mention substantial engineering and computational overhead caused by training and data curation. Parameter efficient fine tuning methods \citep{hu2022lora, dettmers2023qlora, tian2024hydralora} typically trade computational overhead for performance \citep{chen2022revisiting, biderman2024lora}, while still requiring expensive data collection. Additionally, gradient based methods require model weight access, thereby precluding application to powerful closed-source models. In-context approaches, on the other hand, resolve the need for weight access. Retrieval augmented generation \citep{lewis2020retrieval, ram2023context, khandelwal2019generalization} aims to equip the model with new knowledge through similarity based indexing, but is highly sensitive to arbitrary factors such as ordering \citep{liu2024lost}. Our approach builds off of in-context instruction provision, but automatically discovers populations of resources to optimally confer the required knowledge update without any manual, iterative prompt refinement. 

\textbf{Prompting.} Optimal performance in specialized tasks is highly dependent on the prompting strategy \citep{zhou2022least, wang2022self, wang2023plan, madaan2023self}. Recent studies avoid manual prompt engineering by learning prompts through optimizing continuous embeddings \cite{li2024concentrate, sinhababu2025soft, liu2024gpt, qin2021learning, lester2021power}. 
Evolutionary methods have emerged as powerful alternatives for prompt optimization which avoid the need for gradients. \citet{agrawal2025gepa} learn prompts through sampling system-level trajectories and incorporating natural language reflection, while \citet{tong2025evoprompt} prompt LLMs to execute a fixed mutation and crossover process to produce offspring. 
\citet{fernando2023promptbreeder} use an LLM to iteratively and self-referentially mutate prompts while improving the mutation prompts themselves. 
Our method differs from existing evolutionary methods in terms of both method and scope. Regarding method, we depart from LLMs as mutation and crossover operators \citep{tong2025evoprompt, fernando2023promptbreeder}, which are ineffective when the model lacks the knowledge required to meaningfully manipulate task-related information. Instead, we perform mutation with probabilistic draws from the resource pool and crossover with shuffled concatenation, which broadens the exploration space.
Regarding scope, we demonstrate evolution's capability as a search method to accumulate knowledge from diverse sources in unseen domains. 
Where prior work sought to utilize LLMs to iteratively rephrase task queries -- adding instructions to show working, answer in full sentences etc -- we consider the problem setting of evolving contexts from provided corpora for conferring the knowledge necessary to perform \textit{new skills}. 
Hence our approach offers a new avenue towards evolutionary search for novel skill acquisition, moving beyond the evolution of LLM generated outputs towards evolution of knowledge bases themselves.

%% file: Sections/method.tex
\section{Evolutionary context search}

We present \methodacro, a framework that evolves raw text resources into high-utility context for LLM skill acquisition.
This section formalizes the context-evolution problem and details our pipeline, from the initial construction of context units to the specific GA-style operators used to optimize them for task performance.

\subsection{Problem Formulation}
We formalize context optimization as a search problem.
Given text resources $\mathcal{D}$, a target language model $\mathcal{M}$ and a development task set $\mathcal{T}$, we seek an optimal context $C^*$ that maximizes task performance of $\mathcal{M}$.
\begin{equation}
    f(C; M, \mathcal{T})=\mathbf{E}_{(x,y)\in\mathcal{T}}[\mathcal{L}\big(\mathcal{M}(x, C), y\big)]
\end{equation}
where $\mathcal{L}$ is a task-specific scoring metric.
We search for $C^*=\arg \mathrm{max}_{C\subseteq \mathcal{U}}f(C; \mathcal{M}, \mathcal{T})$, where $\mathcal{U} = g(\mathcal{D})$ is a set of context units derived from $\mathcal{D}$, each of which represents an atomic piece of knowledge that can be independently selected, combined, and refined.
A candidate context $C$ is then a structured combination of these units.

\subsection{Algorithm Overview}
\input{Algos/algo1}

\methodacro adapts to diverse tasks by constructing context units in multiple forms (i.e., $g(\mathcal{D})$): raw text from documentation, insights derived from long trajectories, or reusable agentic skills.
Prior evolution-based approaches mutate model-generated content, for example rephrasing self-generated instructions, which inherently limits the search space to knowledge the model can already produce.
By contract, \methodacro draws mutations from provided external text resources, enabling the model to acquire genuinely missing information from its parameters.

Algorithm~\ref{alg:evoctx} presents the complete \methodacro procedure.
The algorithm operates in two phases: initialization and evolution.
During initialization, we construct the initial pool $\mathcal{U}$ from text resources $\mathcal{D}$, extracting units at varying abstraction levels depending on the task setting --- from verbatim source texts to distilled insights and reusable skills (Sec~\ref{subsec:contruct_unit}).
We then sample an initial population $P_0$ of $N$ candidate contexts by drawing units from $\mathcal{U}$ without replacement, ensuring broad coverage of the knowledge pool while maintaining comparable context lengths across candidates.

The evolution loop iteratively improves the population $P_t$.
Each generation evaluates all candidates on the development set $\mathcal{T}$ and selects top performers as elite contexts $P_{\text{elite}}$.
We sample parents fitness-proportionally from $P_{\text{elite}}$ and produce offspring through crossover.
Mutation then introduces variation by adding or replacing units to explore the full unit space $\mathcal{U}$. 
Mutation and crossover thereby jointly expand the population of performant contexts, as determined by those contexts contribution to advancing $\mathcal{M}$ performance on the task.
Finally, LLM-guided refinement resolves logical contradictions inside each offspring.
After $T$ generations, we return the highest-performing context $C^*$.

\subsection{Context Unit Construction}
\label{subsec:contruct_unit}

Different tasks require knowledge at different granularities: code generation benefits from exact syntax and precise documentation, while role-play benefits from abstracted principles.
We therefore design context units at varying abstraction levels, allowing \methodacro to evolve units that span these differing levels of abstraction as necessitated by different tasks. We define three representative types below.


\paragraph{Source texts.} 
These units preserve the precise syntax from source materials, which is vital for replicating exact phrasing or code patterns.
For example, in Domain-Specific Language (DSL) tasks, we include complete code files from NVIDIA's CuTe tutorials, maintaining exact API usage patterns that the model must reproduce faithfully.

\paragraph{Insights.} 
These units distill abstract principles from the provided text source.
Typically, these are actionable rules that capture patterns the model should follow or pitfalls to avoid. 
For example, given failed trajectories from $\tau^2$
-bench \citep{barres2025tau}, we prompt another model to analyze the errors and extracts rules such as ``when processing refunds, retrieve the payment ID directly from the reservation history rather than prompting the user.''

\paragraph{Skills.} 
These units encode reusable procedural knowledge as modular, callable actions. 
Each skill packages domain-specific instructions and multi-step workflows that the model can invoke on demand.
We adapt the Agent Skills format~\citep{anthropic2025skills}, which provides a structured representation for packaging procedural knowledge.
An example is ``write-mha-cutedsl-kernel,'' which encapsulates the procedure for writing multi-head attention kernels in CuTe DSL.
While skills offer a natural structure for procedural knowledge, naively including all available skills can degrade performance due to context distraction. 
\methodacro partially mitigates this by automatically curating task-relevant skills (see Section~\ref{sec:exp}), though improving skill representation and invocation remains an open direction.

\subsection{Evolutionary Operators}

We now detail each evolutionary operator which jointly comprise our context search method. 

\paragraph{Initialization.}
We initialize population $P_0$ of $N$ candidates (typically 32) by sampling units from $\mathcal{U}$ uniformly without replacement.
Each candidate context starts with a predetermined number of units, which we set based on task characteristics: tasks with long units (e.g., complete code files) use fewer units per context, while task with short units (e.g., concise insights) use more.
This maintains comparable context lengths across candidates while preserving diversity among individuals in the initial population.

\paragraph{Fitness Evaluation.}
We score each context $C$ on the development set $\mathcal{T}$ by querying the LLM $\mathcal{M}$ with $C$ as context across all tasks in the development set $\mathcal{T}$.
The fitness $s(C)$ equals the task success rate, normalized to $[0, 1]$.
We use a single rollout per task, which we find sufficient in practice, though additional rollouts could further reduce variance from stochastic $\mathcal{M}$ outputs.

\paragraph{Selection.}
Selection pressure drives the population toward higher-performing contexts. 
We first select the top fraction (e.g., 60\%) of contexts as the elite set $P_{\text{elite}}$. 
We then repeatedly sample parent pairs from this elite set using fitness-proportional selection until we generate $N$ offspring for the next generation. 
For any context $C$, its selection probability $p_C$ equals its fitness normalized by the elite set's total fitness, $p_C = \frac{s(C)}{\sum_{K \in P_{elite}} s(K)}$. 
This focuses reproduction on top performers while maintaining variation within the elite.

\paragraph{Crossover.}
Crossover combines units from two parent contexts $C_a$ and $C_b$ to produce one offspring. 
We concatenate all units from both parents; if the combined set exceeds the maximum context size, we randomly sample from the concatenated context unit pool. 
This allows offspring to inherit complementary information from both parents while respecting context length constraints.

\paragraph{Mutation.}
Mutation introduces variation controlled by a per-context mutation rate (default is 0.1). 
When mutation triggers, we sample a new unit from the full pool $\mathcal{U}$. 
If the context has not reached its maximum size, we add the new unit; otherwise, we replace a randomly selected existing unit. 
This operator ensures the algorithm explores the entire unit space and can escape local optima.

\subsection{LLM-Guided Refinement}
After mutation, an LLM reviews each offspring context to identify and resolve logical contradictions. 
The LLM receives the offspring context and instructions to detect inconsistencies; in principle, failed task examples could also be provided to guide refinement, though we omit this for simplicity in our experiments. 
This refinement step addresses a limitation of blind recombination: merged units may contain conflicting guidance.
The LLM identifies contradictions and either removes or resolves the conflicts.

\vspace{-2mm}



%% file: Algos/algo1.tex
\begin{algorithm}[h]
\caption{Evolutionary Context Search}
\label{alg:evoctx}
\begin{algorithmic}[1]
\REQUIRE Text resources $\mathcal{D}$, dev set $\mathcal{T}$, target model $\mathcal{M}$, population size $N$, generations $T$
\ENSURE Optimal context $C^*$
\STATE $\mathcal{U} \gets g(\mathcal{D})$ \COMMENT{raw text, insights, skills}
\STATE $P_0 \gets \textsc{Initialize}(\mathcal{U}, N)$
\FOR{$t = 0$ \textbf{to} $T-1$}
    \FORALL{$C \in P_t$}
        \STATE $s(C) \gets \textsc{Evaluate}(C, \mathcal{M}, \mathcal{T})$
    \ENDFOR
    \STATE $P_{\text{elite}} \gets \textsc{SelectElite}(P_t, s)$
    \STATE $P_{t+1} \gets \emptyset$
    \FOR{$j = 1$ \textbf{to} $N$}
        \STATE $C_a, C_b \gets \textsc{Sample}(P_{\text{elite}}, s)$
        \STATE $C' \gets \textsc{Crossover}(C_a, C_b)$
        \STATE $C' \gets \textsc{Mutate}(C', \mathcal{U})$
        \STATE $C' \gets \textsc{Refine}(C', \mathcal{T})$ \COMMENT{LLM-guided}
        \STATE $P_{t+1} \gets P_{t+1} \cup \{C'\}$
    \ENDFOR
\ENDFOR
\STATE \textbf{return} $C^* \gets \arg\max_{C \in P_T} \textsc{Evaluate}(C, \mathcal{M}, \mathcal{T})$
\end{algorithmic}
\end{algorithm}

%% file: Sections/experiments.tex
\section{Experiments}
\label{sec:exp}
We empirically validate the benefits of \methodacro in skill acquisition.
We evaluate our method on coding in unseen domain-specific languages with BackendBench~\citep{saroufim2025backendbench} and on multi-turn agentic user assistance with $\tau^2$-Bench~\citep{barres2025tau}, using Gemini-3-Flash-Preview~\citep{googledeepmind2025gemini3flash_modelcard} as the base model for context search.
We also demonstrate cross-model transferability by applying the discovered contexts to Claude-4.5-Sonnet~\citep{anthropic2025claude_sonnet_45_system_card} and DeepSeek-V3.2~\citep{liu2025deepseek}.
In addition, Appendix~\ref{app:sft} demonstrates that SFT remains ineffective for knowledge injection in data-scarce regimes.

\subsection{Experimental Setup}

\paragraph{Baselines.}
We evaluate \methodacro against several baselines.
We implement RAG baselines with LlamaIndex~\citep{Liu_LlamaIndex_2022}, varying two primary axes: chunking strategy and retrieval method.
Chunking strategies include fixed-size splitting (\textbf{Chunk}; 1,024 tokens with 200-token overlap) and \textbf{AST}-based parsing, which segments code at semantic boundaries.
Retrieval methods include \textbf{Dense} (similarity over OpenAI text-embedding-3-small embeddings\citep{openai2024embeddings}), \textbf{BM25} (sparse keyword matching), and \textbf{Hybrid} (reciprocal rank fusion of Dense and BM25).
We sweep $k \in \{5, 10, 20\}$ for Chunk + Dense and find $k=10$ performs best, which we use for all RAG configurations.
Beyond RAG, we include \textbf{Full Context}, which loads all available documentation into the context window, and \textbf{Random Sample}, which randomly selects the same number of context units as \methodacro.

\paragraph{Tasks.}
We evaluate our method on two challenging benchmarks: kernel coding in new Domain-Specific Language (DSL) and multi-turn agentic user assistance.

\textit{BackendBench (CuTeDSL)}~\citep{saroufim2025backendbench} focuses on testing whether models can generate correct GPU kernels in various DSLs (Triton, CUDA, CuTeDSL).
Among these, we choose CuTeDSL to showcase \methodacro capacity to convert newly released textual tutorials into knowledge-rich context capable of guiding the model.
CuTeDSL is NVIDIA's Python DSL built on CUDA Templates for Linear Algebra Subroutines (CUTLASS)~\citep{nvidia2026cutlassdocs}, where text resources are provided by tutorial examples in the repository. 
We randomly select 20 core PyTorch operators, each with 8--12 test cases from PyTorch's operator information (OpInfo) suite.
Appendix~\ref{app:operators} details these operators.
We run 3 evaluations and report the average correctness rate.

\textit{$\tau^2$-Bench (Airline Domain)}~\citep{barres2025tau} evaluates conversational agents on completing user requests through multi-turn interaction, while adhering to domain-specific policies which may conflict with user requests.
The airline domain presents customer service tasks -- booking modifications, cancellations, and policy inquiries -- that must be completed while maintaining airline policy standards.
We obtain text resources by collecting trajectories from GPT-5.2 \citep{openai2025gpt52_systemcard_update} and Gemini-3-Pro \citep{googledeepmind2025gemini3pro_modelcard} on the official training set, then prompting the latter to extract insights from these trajectories. 
Following \citet{barres2025tau}, we use GPT-4.1-2025-04-14~\citep{achiam2023gpt} as the user simulator.
Pass$^k$ for $k \in \{1, 2, 3\}$ measures the rate at which all $k$ trials succeed. 
We evaluate each configuration 3 times with 3 trials, yielding 9 runs total.

\vspace{-2mm}
\paragraph{\methodacro Configuration.}
We run \methodacro for 5 generations (10 for $\tau^2$-Bench) with a population size of 32, selecting the top 60\% as elites with a mutation rate of 0.1.
Each context is limited to a maximum of 10 units, drawn from a pool of 85 source documents for BackendBench and 60 extracted insights for $\tau^2$-Bench.
Fitness is evaluated on 10 development samples per task.
In our experiments, $\mathcal{M}$ is Gemini-3-Flash, and we use Gemini-3-Pro for refinement (Line 15 in Algorithm~\ref{alg:evoctx}).

\subsection{Main Results}

\paragraph{Observation 1: Evolutionary Context Search constructs more effective context than retrieval-based approaches.}
\methodacro consistently outperforms standard retrieval baselines across both the DSL kernel coding and agentic tasks.

\begin{figure}[h]
    \centering
    \includegraphics[width=0.8\linewidth]{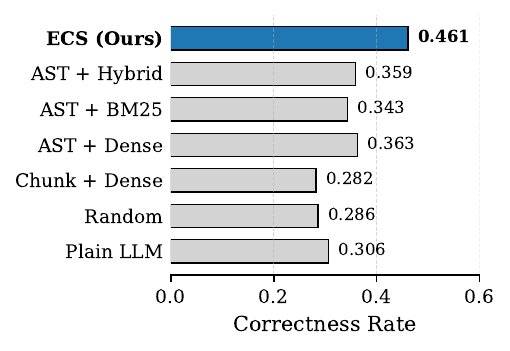}
    \vspace{-3mm}
    \caption{\textbf{Performance comparison on Backend Bench.} Our method (ECS) achieves a 27\% relative improvement over AST + Hybrid. Detailed result are in Appendix~\ref{app:full_beb_result}.
    }
    \label{fig:backend_bench}
\end{figure}

\begin{figure}[h]
    \centering
    \includegraphics[width=0.8\linewidth]{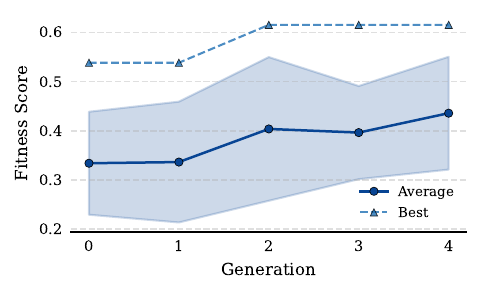}
    \vspace{-3mm}
    \caption{\textbf{Fitness score during the evolutionary search process.}     
    \label{fig:train_fitness}
    }
\vspace{-2mm}
\end{figure}

Regarding BackendBench, where the model must master CuTeDSL using tutorial-based coding resources, Figure~\ref{fig:backend_bench} shows that \methodacro achieves a correctness rate of 0.461, outperforming the strongest retrieval baseline (AST + Dense) by 27\% relative improvement.
We also illustrate the training dynamics in Figure~\ref{fig:train_fitness}.
Beyond this, several patterns emerge. 
First, the chunking strategy is critical: all AST-based methods consistently outperform Chunk + Dense, confirming that preserving semantic boundaries matters for code documentation.
Second, retrieval method choice has limited impact—Dense, Hybrid, and BM25 perform similarly when paired with AST.
Third, naive approaches can hurt: both Chunk + Dense and Random underperform the Plain LLM baseline.
Since Random uses the same context budget as \methodacro, this confirms that arbitrary selection introduces noise, which degrades performance.
This concurs with existing findings that irrelevant context deteriorates RAG's performance relative to the base model \citep{yoran2023making}, implying that RAG is dependent on careful curation.
These results expose a core limitation of retrieval: optimizing chunk-level relevance rather than compositional coherence.
\methodacro addresses this by evolving context combinations that maximize task performance.

This superior performance is further validated on the agentic \textit{$\tau^2$-Bench} in Table~\ref{tab:tau2bench}, where \methodacro achieves 0.767 on Pass$^1$, outperforming the strongest baseline (Full Context at 0.717).
Two observations stand out.
First, the gap widens under stricter metrics: on Pass$^3$, \methodacro maintains 0.683 while Plain LLM and BM25 degrade sharply. 
This indicates that curated contexts enable more consistent policy adherence across repeated trials. 
Second, Full Context outperforms retrieval-based methods, suggesting that broader coverage matters for policy-heavy tasks—yet, \methodacro surpasses Full Context while using far less context, demonstrating that selective curation beats brute-force inclusion.
This reveals a further benefit of \methodacro, which is that our method constructs efficient contexts that confer the required knowledge without context bloat.

\input{Tables/tau2bench}

\vspace{-3mm}
\paragraph{Observation 2: The evolved context is highly transferable across different models, as it encodes semantically meaningful information.}
To assess the generalizability of \methodacro beyond the base model, we evaluate contexts evolved with Gemini-3-Flash on two powerful held-out models: Claude-4.5-Sonnet and DeepSeek-V3.2.

\begin{figure}[t]
    \centering
    \includegraphics[width=0.8\linewidth]{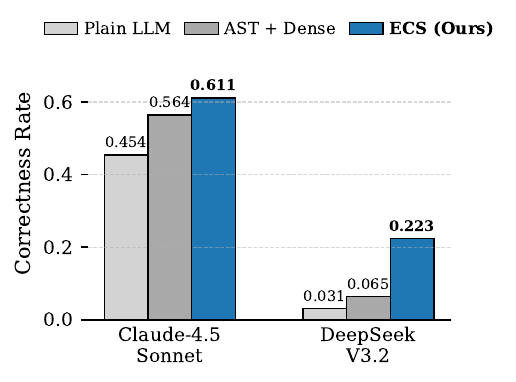}
    \vspace{-3mm}
    \caption{\textbf{Transferability to Unseen Models (BackendBench).} Contexts evolved by \methodacro transfer effectively to models not used during evolution. On DeepSeek-V3.2, where standard retrieval (AST+Dense) fails to provide meaningful gains, \methodacro unlocks significant capability, yielding a significant improvement.}
    \label{fig:transferability}
    \vspace{-5mm}
\end{figure}

As shown in Figure \ref{fig:transferability}, for \textit{BackendBench} \methodacro exhibits strong generalization. With Claude-4.5-Sonnet, the evolved contexts achieve 0.611 correctness rate, surpassing the strong AST+Dense baseline at 0.564. 
Notably, with DeepSeek-V3.2, where standard retrieval fails to provide meaningful gains (0.065 vs 0.031 plain), \methodacro unlocks significant capapbilities, reaching 0.223, a 7x improvement over the plain baseline.
We provide further analysis on this context transfer in Section~\ref{subsec:context_transfer}.


This strong generalization performance holds for \textit{$\tau^2$-Bench} (Table~\ref{tab:tau2_transfer}). 
With Claude-4.5-Sonnet, ECS matches Full on Pass$^1$ and outperforms it on stricter metrics (Pass$^3$: 0.583 vs. 0.500), indicating that \methodacro effectively filters noise that otherwise distracts capable models. 
With DeepSeek-V3.2, ECS improves over Plain and performs close to Full, with significantly fewer tokens (848 vs 5491)
These results demonstrate that \methodacro captures transferable contexts enabling instant skill acquisition—even when evolved with a smaller model (Gemini-3-Flash) and then transferred to larger, more capable models.
Moreover, this result suggests the possibility of using \methodacro as an automated data curation process that extract high-utility data for SFT.

\input{Tables/tau2bench_transfer}
\vspace{-3mm}

\paragraph{Observation 3: Evolutionary Context Search enables effective skill-based augmentation by automatically curating task-relevant skills.}
Agentic skills represent an emerging paradigm for extending LLM capabilities (e.g., Claude Skills), but best practices for skill selection remain an open challenge.
To evaluate \methodacro in this setting, we extract 75 skills from the CuTeDSL documentation, each corresponding to an example code snippet. 
From our experiments, we find that naively including all available skills degrades performance (0.283) compared to the plain LLM baseline (0.306), due to context distraction caused by irrelevant snippets. 
By contrast, \methodacro successfully filters this noise, achieving the highest correctness rate (0.310). 
While this falls short of evolving directly on source text (0.461), it demonstrates that \methodacro can serve as a complementary technique for skill-based systems—transforming noisy skill pools into effective context through evolutionary curation.




%% file: Tables/tau2bench.tex
\begin{table}[h]
    \centering
    \caption{\textbf{Comparison on $\tau^2$-Bench.} ECS significantly outperforms baselines across all Pass metrics.}
    \label{tab:tau2bench}
    \vspace{0.2cm} 
    
    \begin{small} 
    
    \renewcommand{\arraystretch}{1.25} 
    
    \setlength{\tabcolsep}{12pt} 
    
    \begin{tabular}{l ccc} 
        \toprule
        \textbf{\textsc{Method}} & \textbf{Pass$^1$} & \textbf{Pass$^2$} & \textbf{Pass$^3$} \\
        \midrule
        Plain LLM        & 0.622 & 0.489 & 0.400 \\
        Random           & 0.650 & 0.556 & 0.500 \\
        BM25             & 0.641 & 0.550 & 0.483 \\
        Full             & 0.717 & 0.628 & 0.550 \\
        \midrule
        \textbf{ECS (Ours)} & \textbf{0.767} & \textbf{0.710} & \textbf{0.683} \\
        \bottomrule
    \end{tabular}
    \end{small}
\end{table}

%% file: Tables/tau2bench_transfer.tex
    
    

\begin{table}[h]
    \centering
    \caption{\textbf{Transferability on $\tau^2$-Bench.} With Claude-4.5-Sonnet, ECS matches the Full baseline on Pass$^1$ and significantly outperforms it on stricter metrics (Pass$^2$, Pass$^3$). 
    Notably, ECS achieves comparable performance to the Full baselines with substantially fewer tokens, reducing usage from 5491 to 848, a 6$\times$ reduction.
    }
    \label{tab:tau2_transfer}
    \vspace{0.2cm}
    
    \begin{small}
    \renewcommand{\arraystretch}{1.25}
    \setlength{\tabcolsep}{4pt}
    
    \begin{tabular}{l ccc | ccc}
        \toprule
        & \multicolumn{3}{c|}{\textbf{Claude-4.5-Sonnet}} & \multicolumn{3}{c}{\textbf{DeepSeek-V3.2}} \\
        \cmidrule(lr){2-4} \cmidrule(lr){5-7}
        \textbf{Metric} & \textbf{Plain} & \textbf{Full} & \textbf{ECS} & \textbf{Plain} & \textbf{Full} & \textbf{ECS} \\
        \midrule
        Pass$^1$ & 0.600 & 0.678 & \textbf{0.678} & 0.583 & \textbf{0.633} & 0.600 \\
        Pass$^2$ & 0.506 & 0.561 & \textbf{0.617} & 0.478 & \textbf{0.500} & 0.472 \\
        Pass$^3$ & 0.450 & 0.500 & \textbf{0.583} & 0.417 & 0.417 & 0.417 \\
        \midrule
        \# Tokens & 0 & 5491 & 848 & 0 & 5491 & 848 \\
        \bottomrule
    \end{tabular}
    \end{small}
\end{table}

%% file: Sections/analysis.tex
\vspace{-3mm}
\section{Analysis and Ablation}

In this section, we examine the properties of \methodacro in detail.
We first analyze the evolved context qualitatively, then study how models utilize these contexts. 
Finally, we conduct ablation studies and analyze computational costs. 



\vspace{-2mm}
\subsection{Analysis of evolved context.}
Figure~\ref{fig:context-backendbench} presents the context discovered with Gemini Flash from the CuTeDSL code tutorial .
From 85 available documents, \methodacro selected a combination of 7, yet these form a coherent stack spanning multiple abstraction levels. 
\input{Images/context_backendbench}
\
At the kernel layer, \texttt{hopper/fmha.py} provides a comprehensive example of a fused mega-kernel---contributing TMA-based memory transfers, warp-specialized execution, and tensor core MMA(Matrix Multiply-Accumulate) patterns, while also demonstrating the \texttt{cute.math.*} API for arithmetic operations. 
The second kernel example, \texttt{tensorop\_gemm.py}, shows how to construct tiled MMA operations, covering atom layout configuration and threadblock rasterization.
Finally, at the FFI (Foreign Function Interface layer), \texttt{jit\_argument.py} defines C-struct tensor interfaces, enabling data passing between Python and compiled kernels. 
This layered composition suggests that \methodacro does not simply identify isolated code snippets, but in fact accumulates coherent architectural patterns that span from kernel implementation to interfaces.

\vspace{-3mm}
\subsection{Context utilization across models.}
\label{subsec:context_transfer}

\input{Images/context_transfer}

We analyze how exactly the evolved context enables knowledge transfer across models.
Figure~\ref{fig:context-transfer} illustrates one specific example for BackendBench.
The evolved context contains \texttt{fmha.py}, which has the \texttt{cute.math.exp2()} function for softmax operation. 
Although this example never mentions \texttt{atan} or other trigonometric functions, it implicitly teaches the correct API pattern: math operations in CuTeDSL require DSL intrinsics under \texttt{cute.math.*}, not Python's standard library. 
When DeepSeek V3.2 receives this context, it extracts this structural pattern and generalizes it to unseen operators, correctly generating \texttt{cute.math.atan()} despite never observing this specific function in the provided examples.

In contrast, RAG retrieval for the \texttt{atan} operator returns semantically relevant but functionally useless context: GEMM kernels, elementwise addition, and tensor utilities---none of which contain any \texttt{cute.math.*} calls. 
Without working examples, DeepSeek falls back to Python's \texttt{math.atan()}. 
This fails because inside a \texttt{@cute.kernel}, tensor elements are symbolic \texttt{ArithValue} objects, not concrete Python values. 
Only DSL intrinsics, such as \texttt{cute.math.atan()}, are recognized by the JIT compiler and translated to CUDA device code. 
This demonstrates that \methodacro selects context containing transferable structural patterns, whereas RAG retrieves topically similar code that lacks the critical details needed for correct generation.

\vspace{-3mm}
\subsection{Ablation Study}
\vspace{-1mm}

\begin{figure}[h]
    \centering
    \includegraphics[width=0.9\linewidth]{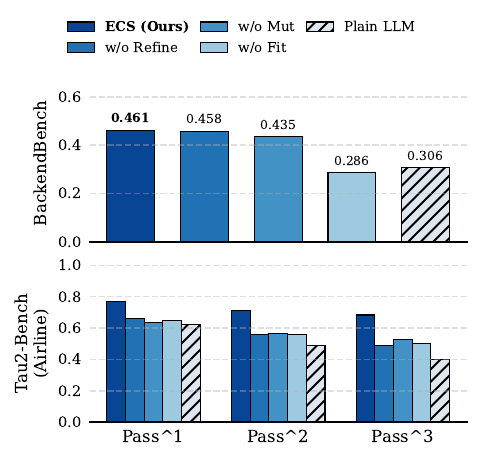}
    \vspace{-3mm} 
    \caption{\textbf{Ablation study.} We evaluate variants by removing components. \textit{Fitness}, \textit{Mutation} are universally critical. \textit{Refinement} is domain-dependent: it is essential for resolving conflicting agentic policies ($\tau^2$-Bench) but has negligible impact on BackendBench.}
    \label{fig:ablation_study}
    \vspace{-6mm}
\end{figure}

To understand the contribution of each component, we evaluate \methodacro variants with fitness-guided selection, mutation, and refinement individually removed.
Figure~\ref{fig:ablation_study} shows the results. 
Fitness-guided selection is critical: removing it causes the largest degradation on both benchmarks, with BackendBench dropping to 0.286 (below Plain LLM). 
Mutation also proves essential, reducing performance on both tasks when removed. 
Interestingly, refinement exhibits domain-dependent behavior: on BackendBench, its removal has negligible impact (0.461 vs 0.458), whereas on $\tau^2$-Bench it causes substantial degradation (Pass$^3$: 0.683 vs 0.488). 
This aligns with our earlier analysis, LLMs struggle to filter tutorial coding samples and tend to retain all of them, while insights often contain explicit contradictions (e.g., conflicting refund rules) that LLM can readily identify. 

\vspace{-3mm}

\subsection{Computational cost.}
\paragraph{Search Cost.}
\methodacro incurs an initial computational cost to discover the optimal context $C^{*}$.
In our experiments, the search budget was limited to $T=10$ generations with a population size of $N=32$, resulting in approximately 320 evaluations on the development set.
While this exceeds the setup cost of standard RAG, which requires computing embeddings for dense retrieval, it is orders of magnitude cheaper than typical SFT or RL workflows, which often necessitate tens of thousands of rollouts to stabilize policy updates~\cite{shao2024deepseekmath}.
Crucially, ECS requires only black-box inference calls, avoiding the substantial memory overhead of gradient updates and optimizer states required by gradient-based methods.

\vspace{-2mm}
\paragraph{Inference Efficiency.}
Once the optimal context is discovered, ECS offers significant efficiency gains over retrieval baselines during deployment.
Because the evolved context $C^{*}$ remains static across all incoming queries for a given task, it is fully compatible with Context Caching (KV-Caching).
In contrast, RAG systems retrieve different context chunks for every unique query, preventing the model from hitting the cached prompt prefix.
Current industry pricing for cached context (e.g. Claude 4.5) charges approximately 10\% of the standard input token price for cache hits~\cite{anthropic_cache}.
Furthermore, caching eliminates the prefill calculation for the context, significantly reducing Time-To-First-Token.
Thus, while ECS requires an upfront search investment, it largely reduces the recurring deployment cost for the context portion compared to dynamic retrieval methods.

%% file: Images/context_backendbench.tex
\begin{figure}[h]
\centering
\small

\textbf{(a) hopper/fmha.py} \texttt{[Kernel]} \hfill \textbf{2,540 lines (58\%)} \\
{\scriptsize \textit{Role: Fused MHA with TMA + TensorCore}}
\vspace{-0.5em} 
\begin{lstlisting}[language=Python, basicstyle=\scriptsize\ttfamily,
  numbers=left, numberstyle=\tiny\color{gray}, xleftmargin=1.5em]
# Q*K^T, softmax, softmax(Q*K^T)*V fused
for j in cutlass.range_constexpr(
    cute.size(acc_qk_mn, mode=[1])):
    acc_qk_mn[i, j] = cute.math.exp2(
        scale_softmax_log2 * acc_qk_mn[i, j] - scale_max,
        fastmath=True)
\end{lstlisting}

\vspace{0.3em}

\textbf{(b) tensorop\_gemm.py} \texttt{[Kernel]} \hfill \textbf{1,012 lines (23\%)} \\
{\scriptsize \textit{Role: Dense GEMM for Ampere architecture}}
\vspace{-0.5em}
\begin{lstlisting}[language=Python, basicstyle=\scriptsize\ttfamily,
  numbers=left, numberstyle=\tiny\color{gray}, xleftmargin=1.5em]
# Creates MMA atom with 16x8x16 shape for MNK
op = cute.nvgpu.warp.MmaF16BF16Op(
    self.ab_dtype, self.acc_dtype, self.mma_inst_shape)
tC = cute.make_layout(self.atom_layout_mnk)
tiled_mma = cute.make_tiled_mma(op, tC, permutation_mnk)
\end{lstlisting}

\vspace{0.3em}

\textbf{(c) jit\_argument.py} \texttt{[FFI]} \hfill \textbf{320 lines (7\%)} \\
{\scriptsize \textit{Role: C-struct tensor interface via LLVM}}
\vspace{-0.5em}
\begin{lstlisting}[language=Python, basicstyle=\scriptsize\ttfamily,
  numbers=left, numberstyle=\tiny\color{gray}, xleftmargin=1.5em]
# FFI: Extract pointer from C-struct
ptr_val = llvm.extractvalue(
    llvm.PointerType.get(), self, [0], loc=loc, ip=ip)
return cute.make_ptr(cutlass.Float32, ptr_val)
\end{lstlisting}

\vspace{0.3em}


\vspace{-0.5em}
\vspace{-3mm}
\caption{\textbf{Core components of the BackendBench evolved context.} The figure illustrates the three most significant code snippets: \textbf{(a)}~A Fused MHA kernel targeting the NVIDIA Hopper architecture (58\% of context); \textbf{(b)}~a dense GEMM kernel configuration for Ampere Tensor Cores (23\%); and \textbf{(c)}~a low-level FFI interface managing LLVM pointer extraction (7\%).}
\vspace{-6mm}
\label{fig:context-backendbench}
\end{figure}

%% file: Images/context_transfer.tex
\begin{figure*}[h]
\centering
\lstdefinestyle{compact}{language=Python, basicstyle=\scriptsize\ttfamily,
  numbers=left, numberstyle=\tiny\color{gray}, xleftmargin=0.8em,
  aboveskip=0pt, belowskip=0pt}
\begin{tabular}{@{}c@{\hspace{0.3em}}c@{\hspace{0.3em}}c@{}}
\textbf{(a) Evolved Context} & \textbf{(b) RAG Baseline} & \textbf{(c) With Evolved Context} \\[-0.2em]
\small\textit{hopper/fmha.py (lines 1372--1384)} & \small\textit{DeepSeek generates wrong API} & \small\textit{DeepSeek generalizes correctly} \\[0.2em]
\begin{minipage}[t]{0.32\textwidth}
\begin{lstlisting}[style=compact, commentstyle=\color{gray}\itshape,
  moredelim={[is][\color{blue}\bfseries]{@}{@}}]
for j in range_constexpr(...):
  acc_qk_mn[i,j] = @cute.math.exp2@(
    scale_softmax_log2 
    * acc_qk_mn[i,j]
\end{lstlisting}
\end{minipage} &
\begin{minipage}[t]{0.32\textwidth}
\begin{lstlisting}[style=compact, commentstyle=\color{red}\itshape,
  moredelim={[is][\color{red}\bfseries]{@}{@}}]
input_val = gA[tidx]
# WRONG: ArithValue is symbolic!
result = @math.atan@(input_val)
\end{lstlisting}
\end{minipage} &
\begin{minipage}[t]{0.32\textwidth}
\begin{lstlisting}[style=compact, commentstyle=\color{teal}\itshape,
  moredelim={[is][\color{teal}\bfseries]{@}{@}}]
input_val = gA[tidx]
# CORRECT: DSL intrinsic
result = @cute.math.atan@(input_val)
\end{lstlisting}
\end{minipage} \\[0.3em]
\colorbox{blue!10}{\small\texttt{cute.math.*} pattern} &
\colorbox{red!10}{\small$\times$ Fails: stdlib on symbolic value} &
\colorbox{green!10}{\small$\checkmark$ Works: DSL intrinsic}
\end{tabular}

\vspace{0.5em}
\small
\begin{tabular}{@{}lccc@{}}
\toprule
\textbf{Operator} & \textbf{RAG Baseline} & \textbf{Evolved Context} & \textbf{Knowledge Transferred} \\
\midrule
\texttt{atan.default}  & 0\% (0/1)  & \textbf{100\%} (1/1)  & \texttt{cute.math.atan()} \\
\texttt{atan2.default} & 0\% (0/9)  & \textbf{88.9\%} (8/9) & \texttt{cute.math.atan2()} \\
\texttt{div.Tensor}    & 0\% (0/18) & \textbf{88.9\%} (16/18) & \texttt{cute.math.*} pattern \\
\midrule
\textbf{Overall (20 ops)} & 6.5\% & \textbf{22.3\%} & $3.4\times$ improvement \\
\bottomrule
\end{tabular}
\caption{\textbf{Cross-model context transfer from Gemini Flash to DeepSeek V3.2 on BackendBench.} 
\textbf{Top:} (a)~Evolved context demonstrates the \texttt{cute.math.*} pattern. 
(b)~Without this context, DeepSeek incorrectly uses Python's \texttt{math.atan()} on symbolic values, causing JIT compilation failures. 
(c)~With evolved context, DeepSeek correctly generalizes to DSL intrinsics. 
\textbf{Bottom:} Quantitative results show evolved context improves pass rate from 6.5\% to 22.3\% ($3.4\times$); e.g., \texttt{div.Tensor} passes 16 out of 18 test cases.}
\vspace{-5mm}
\label{fig:context-transfer}
\end{figure*}

%% file: Sections/conclusion.tex
\vspace{-3mm}
\section{Conclusions}
\vspace{-2mm}

We introduce Evolutionary Context Search (\methodacro), a method that reframes knowledge acquisition as an evolutionary search over text resources.
\methodacro achieves a 27\% relative improvement on BackendBench and 7\% on $\tau^2$-Bench over existing RAG baselines. Furthermore, contexts evolved using Gemini-3-Flash transfer effectively to Claude-4.5-Sonnet and DeepSeek-V3.2, notably yielding a 7× improvement on DeepSeek where standard RAG provides negligible gains.

While \methodacro is highly effective for structured knowledge, its exploration efficiency may be challenged by ``Needle-in-the-haystack'' scenarios within massive corpora.
In such cases, leveraging an LLM for informed population initialization remains a promising refinement.
Most notably, the transferability of our results suggests that \methodacro can serve as an automated data curation mechanism that helps the SFT.
An iterative pipeline, where \methodacro discovers optimal demonstrations and SFT internalizes them, forms a robust paradigm for enabling open-source models to master new skills with minimal human intervention.

%% file: Sections/appendix.tex
\appendix
\onecolumn

\section{BackendBench}

\subsection{Evaluation Operators}
\label{app:operators}

We evaluate our method on 20 PyTorch operators from the BackendBench benchmark, spanning three categories: trigonometric functions, arithmetic operations, and linear algebra primitives. These operators represent a diverse set of computational patterns commonly encountered in deep learning workloads. Table~\ref{tab:operators} provides detailed descriptions of each operator.

\input{Tables_app/backendbench_op20}

\subsection{Full Backend Bench results}
\label{app:full_beb_result}

\noindent
\begin{minipage}{\textwidth}
    \centering
    \makeatletter\def\@captype{table}\makeatother
    \par\vspace{1cm}
    \input{Tables_app/backendbench}
    \par\vspace{2cm}
    
    \input{Tables_app/backendbench_transfer_full}
\end{minipage}

\newpage
\section{Difficulty of Self Supervised Finetuning with Limited Domain Data}
\label{app:sft}

We investigate whether supervised fine-tuning (SFT) could improve open-source model performance on CuTeDSL code generation. 
Our experiments reveal that fine-tuning with limited domain-specific data presents significant challenges.

\subsection{Dataset Construction}
\label{app:sft:dataset}

We curate a training dataset from the CuTeDSL reference documentation, processing 62 Python kernel implementation files and 12 Jupyter notebook tutorials to cover a wide range of GPU architectures. 
To maximize data diversity, we employ three complementary extraction strategies: full-file extraction (50 samples) using module docstrings as queries, full-notebook extraction (12 samples) converted into interleaved markdown and code, and notebook cell extraction (55 samples) pairing individual code cells with their preceding descriptions. 
All samples are standardized into a three-turn chat format compatible with SFT frameworks, consisting of an expert CUDA developer system message, a user query synthesized from the source documentation, and the corresponding ground-truth code.
The dataset statistics are summarized in Table~\ref{tab:sft-dataset}.
\input{Tables_app/sft_data}

\subsection{Experimental Setup and Result}
\label{app:sft:setup}

We fine-tuned the Qwen3-8B model (8.2B parameters) using LoRA (rank=8, $\alpha$=32) applied to all linear layers, yielding 21.8M trainable parameters (0.27\% of the total). 
To optimize performance, we conducted a learning rate sweep across five values spanning three orders of magnitude, with full training configuration details provided in Table~\ref{tab:sft-config}.

\begin{table}[h]
\centering
\caption{SFT Training Configuration}
\label{tab:sft-config}
\begin{tabular}{ll}
\toprule
Hyperparameter & Value \\
\midrule
Base Model & Qwen3-8B \\
Fine-tuning Method & LoRA (rank=8, $\alpha$=32) \\
Trainable Parameters & 21.8M (0.27\%) \\
Learning Rates & \{1e-6, 5e-6, 1e-5, 5e-5, 1e-4\} \\
Global Batch Size & 64 \\
Epochs & 15 \\
Max Sequence Length & 16,384 \\
Precision & bfloat16 \\
\bottomrule
\end{tabular}
\end{table}

Across all learning rates, training failed to converge meaningfully, final loss stagnated at $\sim$5.89 and accuracy remained at baseline levels ($\sim$52\%).
We observed severe gradient instability driven by the data regime rather than hyperparameter selection, with initial gradient norms exceeding $5 \times 10^7$ and values becoming undefined (NaN) by the third epoch. 
Crucially, this lack of convergence resulted in all checkpoints achieving a zero score on the BackendBench task.
These results highlight the fundamental difficulty of fine-tuning on small, specialized datasets with long sequences: with only 117 high-variance examples averaging $\sim$16K tokens, the model lacks sufficient signal to learn generalizable CuTeDSL patterns. 
Consequently, this motivates our shift to a retrieval-augmented in-context learning approach, which resolves the instability of fine-tuning.

%% file: Tables_app/backendbench_op20.tex
\begin{table*}[h]
\centering
\caption{PyTorch operators used in BackendBench evaluation. We evaluate 20 operators across three categories: trigonometric functions (8 ops), arithmetic operations (4 ops), and linear algebra primitives (8 ops).}
\label{tab:operators}
\small
\begin{tabular}{@{}llp{5.8cm}p{5.8cm}@{}}
\toprule
\textbf{Category} & \textbf{Operator} & \textbf{Signature} & \textbf{Description} \\
\midrule
\multicolumn{4}{@{}l}{\textit{Trigonometric Functions}} \\
\addlinespace[0.3em]
& \texttt{acos} & $f: [-1, 1] \rightarrow [0, \pi]$ & Computes element-wise inverse cosine (arccosine) of the input tensor. \\
& \texttt{acosh} & $f: [1, \infty) \rightarrow [0, \infty)$ & Computes element-wise inverse hyperbolic cosine: $\ln(x + \sqrt{x^2 - 1})$. \\
& \texttt{asin} & $f: [-1, 1] \rightarrow [-\frac{\pi}{2}, \frac{\pi}{2}]$ & Computes element-wise inverse sine (arcsine) of the input tensor. \\
& \texttt{asinh} & $f: \mathbb{R} \rightarrow \mathbb{R}$ & Computes element-wise inverse hyperbolic sine: $\ln(x + \sqrt{x^2 + 1})$. \\
& \texttt{atan} & $f: \mathbb{R} \rightarrow (-\frac{\pi}{2}, \frac{\pi}{2})$ & Computes element-wise inverse tangent (arctangent) of the input tensor. \\
& \texttt{atan2} & $f: \mathbb{R}^2 \rightarrow [-\pi, \pi]$ & Computes element-wise two-argument arctangent of $y/x$, using signs to determine quadrant. \\
& \texttt{atanh} & $f: (-1, 1) \rightarrow \mathbb{R}$ & Computes element-wise inverse hyperbolic tangent: $\frac{1}{2}\ln\frac{1+x}{1-x}$. \\
& \texttt{ceil} & $f: \mathbb{R} \rightarrow \mathbb{Z}$ & Rounds each element to the smallest integer greater than or equal to the input. \\
\midrule
\multicolumn{4}{@{}l}{\textit{Arithmetic Operations}} \\
\addlinespace[0.3em]
& \texttt{div} & $(a, b) \mapsto a / b$ & Computes element-wise division. Supports both integer and floating-point semantics. \\
& \texttt{div\_mode} & $(a, b, \text{mode}) \mapsto a / b$ & Element-wise division with explicit rounding: \texttt{trunc} or \texttt{floor} mode. \\
& \texttt{fmod} & $(a, b) \mapsto a - b \cdot \text{trunc}(a/b)$ & C-style remainder (truncated division). Result has same sign as dividend. \\
& \texttt{remainder} & $(a, b) \mapsto a - b \cdot \text{floor}(a/b)$ & Python-style remainder (floored division). Result has same sign as divisor. \\
\midrule
\multicolumn{4}{@{}l}{\textit{Linear Algebra Primitives}} \\
\addlinespace[0.3em]
& \texttt{addmm} & $\beta M + \alpha (A @ B)$ & Matrix-matrix multiply with accumulation. $A \in \mathbb{R}^{n \times m}$, $B \in \mathbb{R}^{m \times p}$, $M \in \mathbb{R}^{n \times p}$. \\
& \texttt{addmv} & $\beta v + \alpha (A @ x)$ & Matrix-vector multiply with accumulation. $A \in \mathbb{R}^{n \times m}$, $x \in \mathbb{R}^{m}$, $v \in \mathbb{R}^{n}$. \\
& \texttt{addbmm} & $\beta M + \alpha \sum_i (A_i @ B_i)$ & Batched matrix multiply with reduction over batch dimension and accumulation. \\
& \texttt{baddbmm} & $\beta M_i + \alpha (A_i @ B_i)$ & Batched matrix multiply with batched accumulation (no reduction). \\
& \texttt{bmm} & $C_i = A_i @ B_i$ & Batched matrix multiplication. $A \in \mathbb{R}^{b \times n \times m}$, $B \in \mathbb{R}^{b \times m \times p}$. \\
& \texttt{dot} & $\sum_i a_i \cdot b_i$ & Dot product of two 1-D tensors $a, b \in \mathbb{R}^{n}$. \\
& \texttt{addr} & $\beta M + \alpha (u \otimes v)$ & Outer product with accumulation. $u \in \mathbb{R}^{n}$, $v \in \mathbb{R}^{m}$, $M \in \mathbb{R}^{n \times m}$. \\
& \texttt{linalg\_cross} & $a \times b$ & Cross product of 3-D vectors, returning vector perpendicular to both inputs. \\
\bottomrule
\end{tabular}
\end{table*}

%% file: Tables_app/backendbench.tex
\caption{\textbf{Comparison between ECS and various RAG baselines in BackendBench}}
\vspace{-3mm}
\small
\centering
\label{tab:backendbench}

\begin{tabular}{@{}l*{7}{r}@{}}
\toprule
\textbf{Operator} & \textbf{Plain} & \textbf{Random} & \textbf{Chunk} & \textbf{AST} & \textbf{AST} & \textbf{AST} & \textbf{ECS} \\
\textbf{} & \textbf{LLM} & \textbf{} & \textbf{+Dense} & \textbf{+Dense} & \textbf{+BM25} & \textbf{+Hybrid} & \textbf{(Ours)} \\
\midrule
acos.default        & 0.333 & 0.444 & 0.444 & \textbf{0.778} & 0.667 & 0.222 & \textbf{0.778} \\
acosh.default       & 0.333 & 0.444 & \textbf{0.778} & 0.667 & 0.667 & 0.667 & \textbf{0.778} \\
addmm.default       & \textbf{0.333} & 0.000 & 0.000 & 0.000 & 0.000 & 0.000 & 0.000 \\
addmv.default       & 0.167 & 0.000 & 0.167 & 0.000 & 0.000 & 0.000 & \textbf{0.222} \\
addbmm.default      & 0.000 & 0.000 & 0.000 & 0.000 & 0.000 & 0.000 & 0.000 \\
baddbmm.default     & 0.000 & 0.333 & 0.333 & 0.333 & 0.000 & 0.667 & \textbf{1.000} \\
dot.default         & 0.000 & 0.000 & 0.000 & 0.000 & 0.000 & 0.000 & 0.000 \\
bmm.default         & \textbf{0.667} & 0.333 & 0.000 & 0.000 & 0.000 & 0.333 & 0.000 \\
addr.default        & 0.111 & 0.167 & 0.000 & \textbf{0.222} & 0.111 & 0.111 & 0.000 \\
asin.default        & \textbf{1.000} & 0.667 & \textbf{1.000} & 0.667 & \textbf{1.000} & \textbf{1.000} & \textbf{1.000} \\
asinh.default       & 0.333 & 0.667 & \textbf{1.000} & \textbf{1.000} & \textbf{1.000} & 0.667 & \textbf{1.000} \\
atan.default        & 0.667 & 0.667 & 0.667 & \textbf{1.000} & 0.667 & 0.667 & \textbf{1.000} \\
atan2.default       & 0.333 & 0.000 & 0.296 & \textbf{0.963} & 0.333 & 0.926 & \textbf{0.963} \\
atanh.default       & 0.667 & \textbf{1.000} & 0.667 & \textbf{1.000} & \textbf{1.000} & \textbf{1.000} & \textbf{1.000} \\
ceil.default        & 0.000 & 0.000 & 0.000 & 0.000 & 0.000 & 0.000 & \textbf{0.333} \\
linalg\_cross.default & \textbf{0.222} & 0.111 & 0.000 & 0.000 & \textbf{0.222} & 0.000 & \textbf{0.222} \\
div.Tensor          & 0.370 & 0.296 & 0.296 & \textbf{0.630} & 0.593 & 0.296 & 0.333 \\
div.Tensor\_mode    & 0.000 & 0.000 & 0.000 & 0.000 & 0.000 & 0.000 & 0.000 \\
fmod.Tensor         & 0.593 & 0.593 & 0.000 & 0.000 & 0.593 & \textbf{0.630} & 0.593 \\
remainder.Tensor    & 0.000 & 0.000 & 0.000 & 0.000 & 0.000 & 0.000 & 0.000 \\
\midrule
\rowcolor{yellow!20}
\textbf{Over Avg Perf} & 0.306 & 0.286 & 0.282 & 0.363 & 0.343 & 0.359 & \textbf{0.461} \\
\bottomrule
\end{tabular}
\vspace{-6mm}

%% file: Tables_app/backendbench_transfer_full.tex
\caption{\textbf{Transferability of evolved contexts generated by ECS across various models on BackendBench}}
\vspace{-3mm}
\small
\centering
\label{tab:transferability}

\begin{tabular}{@{}l*{6}{r}@{}}
\toprule
\textbf{Operator} & \multicolumn{3}{c}{\textbf{Claude-4.5-Sonnet}} & \multicolumn{3}{c}{\textbf{DeepSeekV3.2}} \\
\cmidrule(lr){2-4} \cmidrule(lr){5-7}
\textbf{} & \textbf{Plain} & \textbf{AST} & \textbf{ECS} & \textbf{Plain} & \textbf{AST} & \textbf{ECS} \\
\textbf{} & \textbf{} & \textbf{+Dense} & \textbf{(Ours)} & \textbf{} & \textbf{+Dense} & \textbf{(Ours)} \\
\midrule
acos.default          & 0.111 & 0.667 & \textbf{0.889} & \textbf{0.222} & 0.111 & \textbf{0.222} \\
acosh.default         & 0.444 & \textbf{0.778} & 0.667 & 0.000 & 0.000 & \textbf{0.222} \\
addmm.default         & 0.611 & \textbf{0.667} & 0.389 & \textbf{0.111} & 0.000 & 0.000 \\
addmv.default         & 0.222 & \textbf{0.500} & 0.167 & 0.000 & \textbf{0.167} & \textbf{0.167} \\
addbmm.default        & 0.000 & 0.000 & 0.000 & 0.000 & 0.000 & 0.000 \\
baddbmm.default       & \textbf{0.333} & 0.167 & \textbf{0.333} & 0.000 & 0.000 & 0.000 \\
dot.default           & \textbf{0.667} & 0.333 & \textbf{0.667} & 0.000 & 0.000 & 0.000 \\
bmm.default           & \textbf{1.000} & 0.333 & 0.667 & 0.000 & 0.000 & 0.000 \\
addr.default          & \textbf{0.389} & 0.333 & 0.333 & 0.000 & 0.000 & \textbf{0.222} \\
asin.default          & 0.333 & \textbf{1.000} & \textbf{1.000} & 0.000 & 0.333 & \textbf{0.667} \\
asinh.default         & 0.667 & 0.667 & \textbf{1.000} & 0.000 & \textbf{0.333} & 0.000 \\
atan.default          & 0.333 & \textbf{1.000} & \textbf{1.000} & 0.000 & 0.000 & \textbf{0.667} \\
atan2.default         & 0.630 & 0.926 & \textbf{0.963} & 0.296 & 0.000 & \textbf{0.889} \\
atanh.default         & 0.667 & \textbf{1.000} & \textbf{1.000} & 0.000 & 0.000 & \textbf{1.000} \\
ceil.default          & 0.333 & \textbf{0.667} & \textbf{0.667} & 0.000 & 0.000 & 0.000 \\
linalg\_cross.default & 0.333 & 0.333 & \textbf{0.444} & 0.000 & \textbf{0.111} & \textbf{0.111} \\
div.Tensor            & \textbf{0.778} & 0.593 & 0.370 & 0.000 & 0.185 & \textbf{0.296} \\
div.Tensor\_mode      & 0.000 & \textbf{0.593} & 0.519 & 0.000 & 0.000 & 0.000 \\
fmod.Tensor           & \textbf{0.963} & 0.296 & 0.889 & 0.000 & 0.000 & 0.000 \\
remainder.Tensor      & 0.267 & \textbf{0.433} & 0.267 & 0.000 & \textbf{0.067} & 0.000 \\
\midrule
\rowcolor{yellow!20}
\textbf{Over Avg Perf}& 0.454 & 0.564 & \textbf{0.611} & 0.031 & 0.065 & \textbf{0.223} \\
\bottomrule
\end{tabular}
\vspace{-6mm}

%% file: Tables_app/sft_data.tex
\begin{table}[h]
\centering
\caption{SFT Dataset Statistics}
\label{tab:sft-dataset}
\begin{tabular}{ll}
\toprule
Statistic & Value \\
\midrule
Total training samples & 117 \\
Average sequence length & 15,940 tokens \\
Samples truncated (to 14K tokens) & 24 \\
Source Python files & 62 \\
Source Jupyter notebooks & 12 \\
\midrule
\multicolumn{2}{l}{\textit{By GPU Architecture}} \\
\quad CUDA (generic) & 75 \\
\quad Blackwell & 24 \\
\quad Ampere & 12 \\
\quad Hopper & 3 \\
\quad Multi-GPU/Distributed & 3 \\
\bottomrule
\end{tabular}
\end{table}

%% file: main.bib
@article{liu2024lost,
  title={Lost in the middle: How language models use long contexts},
  author={Liu, Nelson F and Lin, Kevin and Hewitt, John and Paranjape, Ashwin and Bevilacqua, Michele and Petroni, Fabio and Liang, Percy},
  journal={Transactions of the association for computational linguistics},
  volume={12},
  pages={157--173},
  year={2024}
}

@article{lewis2020retrieval,
  title={Retrieval-augmented generation for knowledge-intensive nlp tasks},
  author={Lewis, Patrick and Perez, Ethan and Piktus, Aleksandra and Petroni, Fabio and Karpukhin, Vladimir and Goyal, Naman and K{\"u}ttler, Heinrich and Lewis, Mike and Yih, Wen-tau and Rockt{\"a}schel, Tim and others},
  journal={Advances in neural information processing systems},
  volume={33},
  pages={9459--9474},
  year={2020}
}

@article{khandelwal2019generalization,
  title={Generalization through memorization: Nearest neighbor language models},
  author={Khandelwal, Urvashi and Levy, Omer and Jurafsky, Dan and Zettlemoyer, Luke and Lewis, Mike},
  journal={arXiv preprint arXiv:1911.00172},
  year={2019}
}

@article{ram2023context,
  title={In-context retrieval-augmented language models},
  author={Ram, Ori and Levine, Yoav and Dalmedigos, Itay and Muhlgay, Dor and Shashua, Amnon and Leyton-Brown, Kevin and Shoham, Yoav},
  journal={Transactions of the Association for Computational Linguistics},
  volume={11},
  pages={1316--1331},
  year={2023},
  publisher={MIT Press One Broadway, 12th Floor, Cambridge, Massachusetts 02142, USA~…}
}

@inproceedings{kandpal2023large,
  title={Large language models struggle to learn long-tail knowledge},
  author={Kandpal, Nikhil and Deng, Haikang and Roberts, Adam and Wallace, Eric and Raffel, Colin},
  booktitle={International conference on machine learning},
  pages={15696--15707},
  year={2023},
  organization={PMLR}
}

@inproceedings{li2023large,
  title={Large language models with controllable working memory},
  author={Li, Daliang and Rawat, Ankit Singh and Zaheer, Manzil and Wang, Xin and Lukasik, Michal and Veit, Andreas and Yu, Felix and Kumar, Sanjiv},
  booktitle={Findings of the association for computational linguistics: ACL 2023},
  pages={1774--1793},
  year={2023}
}

@article{petroni2020context,
  title={How context affects language models' factual predictions},
  author={Petroni, Fabio and Lewis, Patrick and Piktus, Aleksandra and Rockt{\"a}schel, Tim and Wu, Yuxiang and Miller, Alexander H and Riedel, Sebastian},
  journal={arXiv preprint arXiv:2005.04611},
  year={2020}
}

@article{hu2022lora,
  title={Lora: Low-rank adaptation of large language models.},
  author={Hu, Edward J and Shen, Yelong and Wallis, Phillip and Allen-Zhu, Zeyuan and Li, Yuanzhi and Wang, Shean and Wang, Lu and Chen, Weizhu and others},
  journal={ICLR},
  volume={1},
  number={2},
  pages={3},
  year={2022}
}

@article{luo2025empirical,
  title={An empirical study of catastrophic forgetting in large language models during continual fine-tuning},
  author={Luo, Yun and Yang, Zhen and Meng, Fandong and Li, Yafu and Zhou, Jie and Zhang, Yue},
  journal={IEEE Transactions on Audio, Speech and Language Processing},
  year={2025},
  publisher={IEEE}
}

@article{shi2025continual,
  title={Continual learning of large language models: A comprehensive survey},
  author={Shi, Haizhou and Xu, Zihao and Wang, Hengyi and Qin, Weiyi and Wang, Wenyuan and Wang, Yibin and Wang, Zifeng and Ebrahimi, Sayna and Wang, Hao},
  journal={ACM Computing Surveys},
  volume={58},
  number={5},
  pages={1--42},
  year={2025},
  publisher={ACM New York, NY}
}

@article{dettmers2023qlora,
  title={Qlora: Efficient finetuning of quantized llms},
  author={Dettmers, Tim and Pagnoni, Artidoro and Holtzman, Ari and Zettlemoyer, Luke},
  journal={Advances in neural information processing systems},
  volume={36},
  pages={10088--10115},
  year={2023}
}

@article{tian2024hydralora,
  title={Hydralora: An asymmetric lora architecture for efficient fine-tuning},
  author={Tian, Chunlin and Shi, Zhan and Guo, Zhijiang and Li, Li and Xu, Cheng-Zhong},
  journal={Advances in Neural Information Processing Systems},
  volume={37},
  pages={9565--9584},
  year={2024}
}

@article{gu2025effectiveness,
  title={On the effectiveness of large language models in domain-specific code generation},
  author={Gu, Xiaodong and Chen, Meng and Lin, Yalan and Hu, Yuhan and Zhang, Hongyu and Wan, Chengcheng and Wei, Zhao and Xu, Yong and Wang, Juhong},
  journal={ACM Transactions on Software Engineering and Methodology},
  volume={34},
  number={3},
  pages={1--22},
  year={2025},
  publisher={ACM New York, NY}
}

@article{li2023unveiling,
  title={Unveiling the pitfalls of knowledge editing for large language models},
  author={Li, Zhoubo and Zhang, Ningyu and Yao, Yunzhi and Wang, Mengru and Chen, Xi and Chen, Huajun},
  journal={arXiv preprint arXiv:2310.02129},
  year={2023}
}

@article{yao2023editing,
  title={Editing large language models: Problems, methods, and opportunities},
  author={Yao, Yunzhi and Wang, Peng and Tian, Bozhong and Cheng, Siyuan and Li, Zhoubo and Deng, Shumin and Chen, Huajun and Zhang, Ningyu},
  journal={arXiv preprint arXiv:2305.13172},
  year={2023}
}

@article{zhong2023mquake,
  title={Mquake: Assessing knowledge editing in language models via multi-hop questions},
  author={Zhong, Zexuan and Wu, Zhengxuan and Manning, Christopher D and Potts, Christopher and Chen, Danqi},
  journal={arXiv preprint arXiv:2305.14795},
  year={2023}
}

@inproceedings{onoe2023can,
  title={Can lms learn new entities from descriptions? challenges in propagating injected knowledge},
  author={Onoe, Yasumasa and Zhang, Michael and Padmanabhan, Shankar and Durrett, Greg and Choi, Eunsol},
  booktitle={Proceedings of the 61st Annual Meeting of the Association for Computational Linguistics (Volume 1: Long Papers)},
  pages={5469--5485},
  year={2023}
}

@article{cottier2024rising,
  title={The rising costs of training frontier AI models},
  author={Cottier, Ben and Rahman, Robi and Fattorini, Loredana and Maslej, Nestor and Besiroglu, Tamay and Owen, David},
  journal={arXiv preprint arXiv:2405.21015},
  year={2024}
}

@article{biderman2024lora,
  title={Lora learns less and forgets less},
  author={Biderman, Dan and Portes, Jacob and Ortiz, Jose Javier Gonzalez and Paul, Mansheej and Greengard, Philip and Jennings, Connor and King, Daniel and Havens, Sam and Chiley, Vitaliy and Frankle, Jonathan and others},
  journal={arXiv preprint arXiv:2405.09673},
  year={2024}
}

@article{chen2022revisiting,
  title={Revisiting parameter-efficient tuning: Are we really there yet?},
  author={Chen, Guanzheng and Liu, Fangyu and Meng, Zaiqiao and Liang, Shangsong},
  journal={arXiv preprint arXiv:2202.07962},
  year={2022}
}

@article{shao2024deepseekmath,
  title={Deepseekmath: Pushing the limits of mathematical reasoning in open language models},
  author={Shao, Zhihong and Wang, Peiyi and Zhu, Qihao and Xu, Runxin and Song, Junxiao and Bi, Xiao and Zhang, Haowei and Zhang, Mingchuan and Li, YK and Wu, Yang and others},
  journal={arXiv preprint arXiv:2402.03300},
  year={2024}
}

@misc{anthropic_cache,
  title        = {Prompt Caching: Reducing Latency and Cost},
  author       = {{Anthropic}},
  year         = {2024},
  howpublished = {\url{https://platform.claude.com/docs/en/build-with-claude/prompt-caching}},
  note         = {Accessed: 2025-01-27. States that cache reads are charged at 0.1x of the base input token price}
}

@article{akkiraju2024facts,
  title={Facts about building retrieval augmented generation-based chatbots},
  author={Akkiraju, Rama and Xu, Anbang and Bora, Deepak and Yu, Tan and An, Lu and Seth, Vishal and Shukla, Aaditya and Gundecha, Pritam and Mehta, Hridhay and Jha, Ashwin and others},
  journal={arXiv preprint arXiv:2407.07858},
  year={2024}
}

@misc{anthropic2025skills,
  author = {{Anthropic}},
  title = {Agent Skills},
  year = {2025},
  howpublished = {\url{https://platform.claude.com/docs/en/agents-and-tools/agent-skills/overview}},
  note = {Accessed: 2025-01-27}
}

@article{zhou2022least,
  title={Least-to-most prompting enables complex reasoning in large language models},
  author={Zhou, Denny and Sch{\"a}rli, Nathanael and Hou, Le and Wei, Jason and Scales, Nathan and Wang, Xuezhi and Schuurmans, Dale and Cui, Claire and Bousquet, Olivier and Le, Quoc and others},
  journal={arXiv preprint arXiv:2205.10625},
  year={2022}
}

@article{wang2023plan,
  title={Plan-and-solve prompting: Improving zero-shot chain-of-thought reasoning by large language models},
  author={Wang, Lei and Xu, Wanyu and Lan, Yihuai and Hu, Zhiqiang and Lan, Yunshi and Lee, Roy Ka-Wei and Lim, Ee-Peng},
  journal={arXiv preprint arXiv:2305.04091},
  year={2023}
}

@article{wang2022self,
  title={Self-consistency improves chain of thought reasoning in language models},
  author={Wang, Xuezhi and Wei, Jason and Schuurmans, Dale and Le, Quoc and Chi, Ed and Narang, Sharan and Chowdhery, Aakanksha and Zhou, Denny},
  journal={arXiv preprint arXiv:2203.11171},
  year={2022}
}

@article{madaan2023self,
  title={Self-refine: Iterative refinement with self-feedback, 2023},
  author={Madaan, Aman and Tandon, Niket and Gupta, Prakhar and Hallinan, Skyler and Gao, Luyu and Wiegreffe, Sarah and Alon, Uri and Dziri, Nouha and Prabhumoye, Shrimai and Yang, Yiming and others},
  journal={URL https://arxiv. org/abs/2303.17651},
  year={2023}
}

@article{liu2024gpt,
  title={GPT understands, too},
  author={Liu, Xiao and Zheng, Yanan and Du, Zhengxiao and Ding, Ming and Qian, Yujie and Yang, Zhilin and Tang, Jie},
  journal={AI Open},
  volume={5},
  pages={208--215},
  year={2024},
  publisher={Elsevier}
}

@article{qin2021learning,
  title={Learning how to ask: Querying LMs with mixtures of soft prompts},
  author={Qin, Guanghui and Eisner, Jason},
  journal={arXiv preprint arXiv:2104.06599},
  year={2021}
}

@article{lester2021power,
  title={The power of scale for parameter-efficient prompt tuning},
  author={Lester, Brian and Al-Rfou, Rami and Constant, Noah},
  journal={arXiv preprint arXiv:2104.08691},
  year={2021}
}

@article{agrawal2025gepa,
  title={Gepa: Reflective prompt evolution can outperform reinforcement learning},
  author={Agrawal, Lakshya A and Tan, Shangyin and Soylu, Dilara and Ziems, Noah and Khare, Rishi and Opsahl-Ong, Krista and Singhvi, Arnav and Shandilya, Herumb and Ryan, Michael J and Jiang, Meng and others},
  journal={arXiv preprint arXiv:2507.19457},
  year={2025}
}

@inproceedings{tong2025evoprompt,
  title={Evoprompt: Evolving prompts for enhanced zero-shot named entity recognition with large language models},
  author={Tong, Zeliang and Ding, Zhuojun and Wei, Wei},
  booktitle={Proceedings of the 31st International Conference on Computational Linguistics},
  pages={5136--5153},
  year={2025}
}

@article{fernando2023promptbreeder,
  title={Promptbreeder: Self-referential self-improvement via prompt evolution},
  author={Fernando, Chrisantha and Banarse, Dylan and Michalewski, Henryk and Osindero, Simon and Rockt{\"a}schel, Tim},
  journal={arXiv preprint arXiv:2309.16797},
  year={2023}
}

@article{li2024concentrate,
  title={Concentrate Attention: Towards Domain-Generalizable Prompt Optimization for Language Models},
  author={Li, Chengzhengxu and Liu, Xiaoming and Zhang, Zhaohan and Wang, Yichen and Liu, Chen and Lan, Yu and Shen, Chao},
  journal={Advances in Neural Information Processing Systems},
  volume={37},
  pages={3391--3420},
  year={2024}
}

@inproceedings{sinhababu2025soft,
  title={Soft Prompt Improves Direct Search Responses},
  author={Sinhababu, Nilanjan and Mitra, Pabitra and Ganguly, Debasis},
  booktitle={Proceedings of the 17th annual meeting of the Forum for Information Retrieval Evaluation},
  pages={79--87},
  year={2025}
}

@misc{barres2025tau,
      title={$\tau^2$-Bench: Evaluating Conversational Agents in a Dual-Control Environment}, 
      author={Victor Barres and Honghua Dong and Soham Ray and Xujie Si and Karthik Narasimhan},
      year={2025},
      eprint={2506.07982},
      archivePrefix={arXiv},
      primaryClass={cs.AI},
      url={https://arxiv.org/abs/2506.07982}, 
}

@article{achiam2023gpt,
  title={Gpt-4 technical report},
  author={Achiam, Josh and Adler, Steven and Agarwal, Sandhini and Ahmad, Lama and Akkaya, Ilge and Aleman, Florencia Leoni and Almeida, Diogo and Altenschmidt, Janko and Altman, Sam and Anadkat, Shyamal and others},
  journal={arXiv preprint arXiv:2303.08774},
  year={2023}
}

@techreport{googledeepmind2025gemini3flash_modelcard,
  title       = {{Gemini 3 Flash} Model Card},
  author      = {{Gemini Team}},
  institution = {Google DeepMind},
  type        = {Model Card},
  year        = {2025},
  month       = dec,
  url         = {https://storage.googleapis.com/deepmind-media/Model-Cards/Gemini-3-Flash-Model-Card.pdf},
  note        = {Published December 2025},
}

@article{liu2025deepseek,
  title={Deepseek-v3. 2: Pushing the frontier of open large language models},
  author={Liu, Aixin and Mei, Aoxue and Lin, Bangcai and Xue, Bing and Wang, Bingxuan and Xu, Bingzheng and Wu, Bochao and Zhang, Bowei and Lin, Chaofan and Dong, Chen and others},
  journal={arXiv preprint arXiv:2512.02556},
  year={2025}
}

@techreport{anthropic2025claude_sonnet_45_system_card,
  title       = {Claude Sonnet 4.5 System Card},
  author      = {{Anthropic}},
  institution = {Anthropic},
  type        = {System Card},
  year        = {2025},
  month       = sep,
  url         = {https://www-cdn.anthropic.com/963373e433e489a87a10c823c52a0a013e9172dd.pdf},
}

@techreport{openai2025gpt52_systemcard_update,
  title       = {Update to GPT-5 System Card: GPT-5.2},
  author      = {{OpenAI}},
  institution = {OpenAI},
  type        = {System Card},
  year        = {2025},
  month       = dec,
  url         = {https://cdn.openai.com/pdf/3a4153c8-c748-4b71-8e31-aecbde944f8d/oai_5_2_system-card.pdf},
  note        = {Dated December 11, 2025},
}

@techreport{googledeepmind2025gemini3pro_modelcard,
  title       = {{Gemini 3 Pro} Model Card},
  author      = {{Gemini Team}},
  institution = {Google DeepMind},
  type        = {Model Card},
  year        = {2025},
  month       = nov,
  url         = {https://storage.googleapis.com/deepmind-media/Model-Cards/Gemini-3-Pro-Model-Card.pdf},
  note        = {Model card published November 2025},
}

@software{saroufim2025backendbench,
  author = {Mark Saroufim and Jiannan Wang and Bert Maher and Sahan Paliskara and Laura Wang and Shahin Sefati and Manuel Candales},
  title = {BackendBench: An Evaluation Suite for Testing How Well LLMs and Humans Can Write PyTorch Backends},
  year = {2025},
  url = {https://github.com/meta-pytorch/BackendBench}
}

@software{Liu_LlamaIndex_2022,
author = {Liu, Jerry},
doi = {10.5281/zenodo.1234},
month = {11},
title = {{LlamaIndex}},
url = {https://github.com/jerryjliu/llama_index},
year = {2022}
}

@manual{nvidia2026cutlassdocs,
  title        = {NVIDIA CUTLASS Documentation (CUTLASS 4.3.5)},
  author       = {{NVIDIA}},
  organization = {NVIDIA},
  year         = {2026},
  month        = jan,
  url          = {https://docs.nvidia.com/cutlass/latest/},
  note         = {Online documentation for CUTLASS 4.3.5 (Jan 2026); accessed 2026-01-28},
}

@article{yoran2023making,
  title={Making retrieval-augmented language models robust to irrelevant context},
  author={Yoran, Ori and Wolfson, Tomer and Ram, Ori and Berant, Jonathan},
  journal={arXiv preprint arXiv:2310.01558},
  year={2023}
}

@misc{openai2024embeddings,
  title        = {New Embedding Models and API Updates},
  author       = {OpenAI},
  year         = {2024},
  howpublished = {\url{https://openai.com/index/new-embedding-models-and-api-updates}},
  note         = {Announcing text-embedding-3-small and text-embedding-3-large}
}

@article{katoch2021review,
  title={A review on genetic algorithm: past, present, and future},
  author={Katoch, Sourabh and Chauhan, Sumit Singh and Kumar, Vijay},
  journal={Multimedia tools and applications},
  volume={80},
  number={5},
  pages={8091--8126},
  year={2021},
  publisher={Springer}
}
